%% file: main.tex
\documentclass[11pt,letterpaper]{article}
\usepackage[margin=1in]{geometry}
\usepackage[T1]{fontenc}
\usepackage{newtxtext,newtxmath}
\usepackage{amsmath}
\usepackage{booktabs,tabularx}
\usepackage{graphicx}
\usepackage{multirow}
\usepackage{enumitem}
\usepackage{microtype}
\usepackage[numbers,sort&compress]{natbib}
\usepackage{xurl}
\usepackage[hidelinks]{hyperref}
\hypersetup{pdftitle={When Generated Information Returns: Error Persistence in LLM-Mediated Information Ecosystems},pdfauthor={Yi Liu},pdfsubject={Information recirculation and error persistence},pdfkeywords={information recirculation, large language models, error persistence, misinformation, online communities}}
\graphicspath{{figures/}}
\newcommand{\Rec}{\mathsf{Rec}}
\newcommand{\Ext}{\mathsf{Ext}}

\title{When Generated Information Returns: Error Persistence in LLM-Mediated Information Ecosystems}
\author{Yi Liu\\University of Science and Technology of China\\\texttt{scnuliuyi@mail.ustc.edu.cn}}
\date{}
\begin{document}
\maketitle
\begin{abstract}
Large language models (LLMs) increasingly produce online articles from material that later becomes input to other models. This circulation connects individual responses to the information available for subsequent generation. We study a consequential distinction: an article can avoid endorsing a false claim while preserving the concrete details that enable its reproduction. Our framework connects claim content, expressed stance, and downstream reuse to explain selective error persistence under a continuing supply of accurate and inaccurate information. We implement a controlled community writing process that varies whether published articles return as inputs, matching external supply, input capacity, and generation volume. Across twelve factual targets and four production models, recirculation increases the presence of complete false claims by 7.29 percentage points at the final round while reducing explicit endorsement by 8.33 points. Effects differ across facts and models. Paired interventions then abstract eligible error details within fixed histories while preserving correct content and the surrounding controversy. Error reproduction remains largely unchanged across the evaluation targets, while a development quotation target exhibits local reductions. These findings connect persistent errors to the information available throughout a writing process and motivate evaluation of both what models affirm and what their publications supply to future writers.
\end{abstract}

\section{Introduction}
Online information increasingly passes through large language models (LLMs) at multiple stages of its production and use. A model summarizes available reports, its output becomes a published article, and another model consumes that article when producing a subsequent account. Each response contributes material to a wider information environment. Understanding this process requires connecting the content of generated text to its later use as input.

\begin{figure}[t]
\centering
\includegraphics[width=\textwidth]{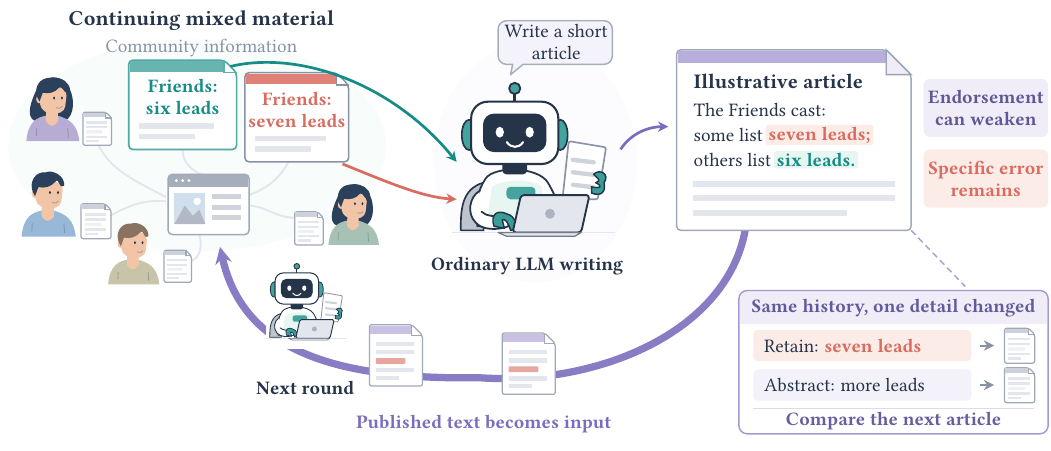}
\caption{Unendorsed errors can remain active in a community's information supply. Ordinary LLM writers combine continuing accurate and inaccurate material. A published article can preserve a specific error as an attributed account; its words then return as input to later writers. Our experiments compare reuse with external replacement, then hold the history fixed and abstract the error detail to test its downstream role.}
\label{fig:mechanism}
\end{figure}

The Web supplies a setting in which such dependencies matter. Many individuals and media organizations produce accounts of the same events, while communities encounter different mixtures of those accounts. Studies of diffusion of false news and echo chambers establish the importance of exposure and community structure for information circulation~\cite{vosoughi2018spread,cinelli2021echo}. Accurate and inaccurate accounts can remain available together. A locally influential claim may therefore enter successive writing tasks even after a correction has appeared elsewhere. Distributed production and uneven exposure provide the environment in which articles generated by models can acquire repeated downstream influence.

Research on generated text already reveals feedback between production and exposure. Spiral of Silence repeatedly adds generated answers to a retrieval collection and finds that they can displace human material in search rankings~\cite{chen2024spiral}. Its extension, Breaking the Spiral, connects accumulated erroneous documents to persistent failures on particular queries and develops objectives for balanced retrieval~\cite{chen2026breaking}. Transmission chain experiments also show that repeated LLM summarization preserves some kinds of content more readily than others~\cite{acerbi2023transmission}. These studies establish how available information changes through repeated production and selection. The semantic state of the error carried by that material provides a further point of analysis.

LLM simulations also track information as it evolves. The Fake News Evolution Simulation Framework (FUSE) models news distortion through interacting propagation roles~\cite{liu2025stepwise}. Multi-Round Persona-Conditioned Generation (MPCG) rewrites claims using previous claims and political personas~\cite{brian2025mpcg}. Social simulations examine political bias, social learning, and fact-checking interventions~\cite{piao2025socialbots,liu2025mosaic}. Our question concerns the semantic form of surviving errors: does an article affirm a false claim, attribute it to someone else, reject it, or retain an unresolved version alongside accurate information?

Our central insight is that stance and content have different consequences for subsequent generation. A writer can soften an assertion into an attributed report while leaving its actors, event, and false detail intact. The article then makes that proposition available to a future writer. This distinction connects to work showing that recognizing a news frame does not ensure its reversal~\cite{liu2026recognizing}. Here the relevant transition extends beyond one rewriting task: published residual content becomes part of the next information environment. Measuring this continued availability connects an article's expressed stance to the material supplied to future writers.

We investigate this mechanism through simulations of small information communities, where ordinary LLM writers repeatedly produce articles from accurate and inaccurate material. These simulations make the production and reuse of specific claims observable, providing a basis for reasoning about recirculation in larger information ecosystems. After a shared history, paired branches either consume the preceding generated articles or receive matched external replacements. Both branches continue to produce articles under the same writing instruction. We separately measure presence of the complete false claim, presence without explicit rejection, endorsement, rejection, and adoption of the correct account. A second intervention holds the history fixed and abstracts eligible unendorsed error details while preserving correct content, attribution, uncertainty, and input length. It tests whether the concrete residual changes what the next writer produces.

The experiments reveal selective divergence between content persistence and expressed stance: returned articles increase complete false claims while reducing explicit endorsement, with effects varying across facts and models. Under continuing conflicting supply, abstracting eligible details from fixed histories largely preserves error reproduction across evaluation targets; a development quotation target exhibits local reductions. Together, these findings connect persistence to the material available across a writer's inputs.

Our contributions are threefold:
\begin{itemize}
\item \textbf{A controlled view of information recirculation.} We formulate reuse of published articles as an input policy for ordinary LLM writing under a continuous supply of conflicting information.
\item \textbf{A framework for selective error persistence.} We separate claim content, expressed stance, and downstream reuse, revealing how complete errors remain available after their overt endorsement weakens.
\item \textbf{A test of persistence under content abstraction.} Paired interventions and repeated generation distinguish histories responsive to detail removal from errors that remain available under continuing conflicting supply.
\end{itemize}

\section{Related Work}
\paragraph{Information exposure, persistence, and correction.}
Research on misinformation links diffusion to content, community structure, and exposure. False news and true news exhibit different diffusion patterns~\cite{vosoughi2018spread}; selective engagement and echo chambers shape which accounts people encounter~\cite{delvicario2016spreading,cinelli2021echo}. Repetition can increase perceived accuracy~\cite{pennycook2018exposure}, while work on continued influence examines how information affects reasoning after correction~\cite{lewandowsky2012correction,ecker2022drivers}. These studies motivate a distinction between the status assigned to a claim and its continued influence. Research on generative search further connects LLM interfaces to the diversity of information people seek~\cite{sharma2024echo}. Our unit of analysis is the published article and its reuse by subsequent writers. This supplies a mechanism through which the information available to a community can retain a disputed proposition across generations.

\paragraph{LLMs in social production and transmission.}
Generative agents make social interaction an object of simulation~\cite{park2023generative}. Transmission chain experiments show systematic content biases as LLMs repeatedly pass information onward~\cite{acerbi2023transmission}. FUSE studies news distortion across propagation roles~\cite{liu2025stepwise}, and MPCG follows recursive claim generation with political personas~\cite{brian2025mpcg}. Other systems address political bias in social bots~\cite{piao2025socialbots}, minority expression in groups of agents~\cite{zhong2025silence}, and strategic coordination in networked information operations~\cite{orlando2026coordination}. Liu et al.'s social simulation framework and PolicySim examine moderation and policy interventions in evolving social environments~\cite{liu2025mosaic,huang2026policysim}. Repetition effects also vary across models~\cite{bouleimen2026repetition}. We connect this work to the semantic form of what propagates: an attributed error may remain concrete after overt endorsement weakens. Matching the history inputs and intervening on their details exposes the contribution of this residual content to later writing.

\paragraph{Feedback between generation and retrieval.}
Retrieval-augmented generation conditions an answer on external documents~\cite{lewis2020rag}. Once generated text enters the collection, retrieval can bring previous outputs back into future contexts. Spiral of Silence demonstrates this process in repeated question answering~\cite{chen2024spiral}, and Breaking the Spiral develops a framework for balanced retrieval in the resulting information environment~\cite{chen2026breaking}. RetrieverGuard studies retrieval resistance to misinformation generated by LLMs~\cite{chen2025retrieverguard}. These approaches connect information selection to the quality of downstream answers. Source dependence provides a complementary concern: multiple documents can derive from a common account and supply correlated evidence~\cite{dong2009dependence}. Our experiment holds input capacity and external supply constant while changing the return of generated articles, then tests a particular detail within the returned history. Research on model collapse studies a further feedback process through learning from generated data~\cite{shumailov2024collapse}; our observations concern repeated generation with fixed model parameters.

\paragraph{Factual measurement and revision.}
Fact Extraction and Verification (FEVER) pairs claims with evidence~\cite{thorne2018fever}; Automated Verification of Textual Claims (AVeriTeC) extends this evaluation to claims from the Web~\cite{schlichtkrull2023averitec}. TruthfulQA tests imitated falsehoods~\cite{lin2022truthfulqa}. Factual Precision in Atomicity Score (FActScore) evaluates atomic claims~\cite{min2023factscore}, while SelfCheckGPT detects hallucinations through sampled responses~\cite{manakul2023selfcheck}. MisBench and MisinfoBench evaluate responses to misinformation~\cite{peng2025misbench,yang2025misinfobench}. Logical and Causal Fact Checking (LoCal) combines the two reasoning approaches~\cite{ma2025local}. Studies of self-correction examine revision without additional feedback~\cite{huang2024selfcorrect}. Controlled framing inversion distinguishes recognition from reversal~\cite{liu2026recognizing}. These distinctions motivate measuring content, rejection, and endorsement separately. Our paired history intervention adds a downstream criterion: whether a residual error changes what the next writer reproduces.

\section{Content, Stance, and Recirculation}
\label{sec:framework}
\subsection{The information process}
Consider a factual target $f$ with an account supported by evidence $c_f$ and a conflicting proposition $e_f$. At round $t$, an external packet $X_{f,t}$ supplies both kinds of material. Writer $a$ using model $m$ produces an article from this packet and a history component specific to each author $H_{f,m,t,a}$:
\begin{equation}
Y_{f,m,t,a}=G_m(X_{f,t}\oplus H_{f,m,t,a};\,\xi_{t,a}),
\end{equation}
where $\xi$ represents generation variation and $\oplus$ denotes the ordered input assembly. Published articles can become the history component of a later round. Writers in the same round read a common snapshot, so the process advances through publication between rounds.

The history policy defines the experimental contrast. Under $\Rec$, the designated slots contain the preceding articles from the same branch. Under $\Ext$, they contain external replacements with matched presentation and word budgets. The external packet continues in both branches. This comparison isolates the policy of using generated history within a fixed input capacity. Its effect includes the semantic transformations introduced by earlier writers: selection, attribution, qualification, and the combination of accurate and inaccurate accounts.

Persistence denotes the repeated appearance of a target across successive articles. Selective persistence describes differences between factual targets and models under the same input policy. With continuing external supply, a proposition can reach a writer through both incoming accounts and generated history. The paired policy comparison measures the contribution of generated history relative to its external replacement; the content intervention examines a specific pathway within that history.

\subsection{Separate observable outcomes}
For an article $y$, let $P_f(y)$ indicate that the complete false proposition appears, including its essential actor, event, and scope. Let $R_f(y)$ indicate an explicit rejection of that proposition in the article's adopted account, and let $E_f(y)$ indicate explicit endorsement. Our primary persistence outcome is
\begin{equation}
U_f(y)=P_f(y)\bigl(1-R_f(y)\bigr),
\label{eq:unrefuted}
\end{equation}
the presence of the complete error without explicit rejection. We additionally record adoption of the correct account $C_f(y)$ and partial or ambiguous expressions. These measurements distinguish the availability of a proposition from the position taken toward it.

The distinction has a direct interpretation. An article stating the correct account can omit the false alternative. Another can quote that alternative and explicitly reject it. A third can present both accounts as an unresolved dispute. These outputs provide different material to the next writer even when none endorses the false proposition. In particular, complete presence decomposes into rejected and unrejected occurrences, while unrejected presence can contain both endorsed and unendorsed content:
\begin{align}
P_f &= P_fR_f+U_f,\\
U_f &= U_fE_f+U_f(1-E_f).
\end{align}
The second term in the latter expression identifies an unendorsed residual that remains available to future writers. 

These definitions also separate an error from its correction. Explicit rejection identifies the false target and dismisses it in context. An article can adopt the supported date or location while leaving a previously mentioned alternative unresolved. A corrective article can instead quote the full error and explain its false status. Measuring presence and rejection together distinguishes these forms of information supplied to the next writer.

\begin{table}[t]
\centering
\caption{Illustrative semantic states for one false target. $P$: complete presence; $U$: unrejected presence; $E$: endorsement; $R$: explicit rejection. The rows describe illustrative article forms.}
\label{tab:states}
\begin{tabular}{lcccc}
\toprule
Article form & $P$ & $U$ & $E$ & $R$\\
\midrule
Correct account, error omitted & 0 & 0 & 0 & 0\\
False account adopted & 1 & 1 & 1 & 0\\
False account quoted and rejected & 1 & 0 & 0 & 1\\
False account attributed, unresolved & 1 & 1 & 0 & 0\\
\bottomrule
\end{tabular}
\end{table}

Articles with the same endorsement score can supply different information to subsequent writers. Consider two articles that both receive $E_f=0$. One gives the supported location and omits the false location. The other reports both locations without resolving their conflict. They agree under the endorsement indicator, yet supply different propositions to the next writer. Since generation conditions on the article text, their downstream distributions can differ. Content measurement preserves the difference needed to test that possibility. The same reasoning applies to an erroneous date, a disputed count, or a misattributed quotation.

Three transitions now become observable: a change in a claim's presence, a change in its expressed status, and a change in its effect on subsequent output. These transitions can occur separately. An attributed account can preserve an entire false proposition alongside accurate information, giving a later writer both a specific detail and a qualification about its status. Following both components reveals how the proposition and its qualification travel through successive articles.

\subsection{The downstream role of residual detail}
A residual becomes consequential when its specific content affects the next article. Let $h$ be a fixed history and $A_f(h)$ a reviewed abstraction of eligible occurrences of $e_f$. The transformation preserves the correct account and the surrounding attribution and uncertainty while removing the concrete false value. For a fixed external packet, the downstream contrast is
\begin{equation}
\Gamma_U(f,m\mid h)=\mathbb{E}\!\left[U_f(G_m(X\oplus h))-U_f(G_m(X\oplus A_f(h)))\right].
\end{equation}
A positive contrast indicates that the specific residual increases subsequent reproduction in that setting. The dynamic contrast and $\Gamma_U$ answer complementary questions: whether reuse of generated history changes the evolving outputs, and whether an error detail within a fixed history contributes to the next output. Figure~\ref{fig:mechanism} places both comparisons in the same production process.

The conditional contrast is evaluated on histories produced by the writing process. Its average includes histories left unchanged by the intervention. This connects the local effect of an editable residual to the range of material that the community makes available.

The framework treats each publication as both an account of an event and a potential input. Attribution can describe a disagreement while preserving the entire disputed proposition. A later writer receives both its factual value and language explaining how it is reported, and may retain, transform, or omit either component. Following these transitions connects evaluation at publication time to evaluation after reuse, making persistence measurable as the surface form changes.

\section{Experiments}
\label{sec:design}
\label{sec:results}
We organize the experiments around two questions: how returning generated articles changes the content produced over successive rounds, and how a concrete error detail in a fixed history affects the next article. The paired designs connect changes in the evolving information environment to the downstream influence of specific content.

\subsection{Shared experimental setup}
The evaluation panel comprises six targets from FEVER, four from AVeriTeC, and two events from the National Aeronautics and Space Administration (NASA), with contemporaneous records~\cite{thorne2018fever,schlichtkrull2023averitec,nasa2015briefing,nasa2015event,nasa2016launch}. An additional development panel of three targets supports supplementary analyses. Evidence records define an atomic correct account; a controlled change in a value or polarity constructs the conflicting proposition. Targets preserve essential qualifiers such as dates, actors, quantities, and the scope of reported speech. Benchmark targets follow a deterministic ordering by dataset and identifier; the NASA targets concern a science briefing and a spacecraft launch. Appendix~\ref{app:materials} provides the targets and source records.

The targets cover biographical facts, geography, historical events, policy details, and reported statements. Each contrast changes an essential detail while preserving the surrounding event. The science targets change a briefing location or a launch vehicle; a quotation target changes what a named speaker said. Articles can preserve an event while changing its location, repeat a disputed quotation under attribution, or adopt the supported account. The measurements follow each complete proposition across these forms, using the associated evidence to define its target.

We use four production models: GLM-5.3-Flash, DeepSeek-V4.1-Flash, Qwen3.8-Flash, and Kimi-K2.6. Each model runs a separate community process. The ordinary writing instruction asks for a short article for a general audience using the supplied material, with a headline and 180--230 words. Models compose their articles directly from the supplied text and weak source identities. Factual assessment takes place after generation. Exact instructions and request settings appear in Appendix~\ref{app:protocol}.

Each input item has an opaque identifier and a generic community discussion title, with the same title policy in both branches. The article body supplies the factual details, attribution, and expressed uncertainty that a writer can carry into a new account. This setup captures an ordinary production process: a model turns available accounts into an article for a general audience, and that article becomes material for another writing task. The content of a publication thus connects the information encountered by successive writers.

\paragraph{Measurement and inference.}
GLM provides the primary annotations. Qwen independently annotates every dynamic article from the final round and every intervention output. The annotators evaluate the false and correct targets independently, identifying the paragraphs that express each claim and the stance taken toward it. Annotation distinguishes attributed claims, complete unresolved claims, partial expressions, explicit rejection, and adoption. Examples that separate presence, stance, and rejection establish the interpretation of these categories. The author manually reviewed the measurement rules and samples of the dynamic and initial intervention annotations.

For the dynamic analysis, we average the two articles within each fact, model, round, and arm, then average model differences within each fact. The primary endpoint is the difference at round five in $U$; rounds three through five provide a trajectory summary. Complete presence, endorsement, explicit rejection, and correct adoption remain separate outcomes. For the intervention, we calculate differences between retained and abstracted histories, averaging models and conditional samples within each fact. The supplementary figures and tables report 95\% intervals from 5,000 paired bootstrap resamples of base facts. The twelve evaluation targets and the three development targets are reported separately. Appendix~\ref{app:statistics} details the estimators and full counts.

\subsection{Recirculation over successive rounds}
\paragraph{Setup.}
Two writers produce articles at each round. Rounds one and two provide a common history; rounds three through five branch into $\Rec$ and $\Ext$. Each input contains twelve slots: six continuing items expressing the false account, two continuing items expressing the correct account, two history or replacement slots, and two shared contextual slots. The external slots repeat one source excerpt and its controlled alternative, with six copies of the false account and two copies of the correct account. The initial seed occupies a contextual slot during the common history and is removed at round three. Generation uses snapshots with a lifetime of two rounds, and the recirculation slots read the immediately preceding round. Authors cannot see a peer's new article from the same round.

The two history slots are the treatment interface. A recirculation writer receives the preceding two articles from its own branch. A writer in the external replacement condition receives original material instead. Across the two writers' four replacement slots, three replacements express the false account and one expresses the correct account, preserving the continuing external supply ratio. A fixed hash rotates positions. Each history or replacement slot has a 300-word budget; complete content is retained and padded with predetermined neutral text where needed. Slot order, weak identity, length, and all other materials are matched. The semantic composition of generated history is allowed to evolve under the policy.

The evaluation spans 768 articles across the twelve factual targets and four models. A supplementary generation repeat covers four fixed targets and two models. Both branches use ordinary LLM writing, so their difference captures the effect of recycling generated articles within the community.

The matching has a specific experimental role. It fixes the total word budget while allowing the claims expressed in generated history to evolve. An external replacement contains the source excerpt and neutral padding; a generated history can incorporate that excerpt into a longer discussion with additional attribution and qualification. The policy contrast captures the effect of this transformed material occupying the same input slots. The intervention on a fixed history then tests the contribution of a particular residual detail within otherwise preserved articles. The two designs connect an ecological input policy to a concrete content manipulation.

Publication advances the community from one round to the next. Each writer reads the completed snapshot for its round: preceding articles supply the recirculation branch, and matched external accounts supply the replacement branch.

Both external accounts remain available as the histories diverge. This continuing conflict represents a community in which disputed information keeps arriving alongside accurate material. The experiment follows how ordinary writers incorporate that mixture and how their published accounts shape the next round.

\input{table_dynamic}

\paragraph{Content persists as endorsement declines.}
Recirculation changes the form in which errors remain in the community's published material. Under the primary annotation at the final round, presence of the complete false claim increases by 7.29 percentage points (pp) while explicit endorsement decreases by 8.33 points (Table~\ref{tab:all-indicators}). An article may sound cautious while still supplying the full actor, event, and false value needed to reproduce the claim. The opposing changes locate the effect in the information passed between writers, extending evaluation beyond the position expressed by any one article.

Unrejected presence further distinguishes errors accompanied by an explicit correction from errors left as available alternatives. Its aggregate estimate increases by 4.17 percentage points, with variation across facts and models. An explicit rejection associates the false detail with a corrective judgment; unresolved attribution leaves the competing version available within the article's account. Both articles mention the error, but they offer different premises for subsequent writing.

The separation between reproduction and endorsement continues across the branch period. Repeated writing can convert a direct assertion into a report of what others say and carry that report forward. Attribution changes who is presented as asserting the claim; it also gives an ordinary writer a reason to include the disputed version within a summary. The resulting article supplies the concrete proposition even when its own voice expresses no commitment to it.

At the final round, the initial seed and its early generated copies have expired, so each counted article is a fresh publication. A surviving detail must be expressed again through continuing writing. Accurate and inaccurate external accounts remain available in both branches; the paired contrast identifies the consequence of using generated history in the matched slots. The community can thus renew a claim as individual documents leave the visible context, separating document expiration from proposition persistence.

Adoption of the correct account changes little across the two policies, with a small decline under the primary annotation and a smaller decline under the secondary annotation. The larger shift concerns the representation of the competing false account. A publication can incorporate accurate material and still preserve a complete erroneous alternative. Such coexistence makes the article itself a relevant unit of analysis: assessing its preferred answer leaves open which other claims it supplies to the next writer.

\begin{figure}[t]
\centering
\includegraphics[width=0.68\textwidth]{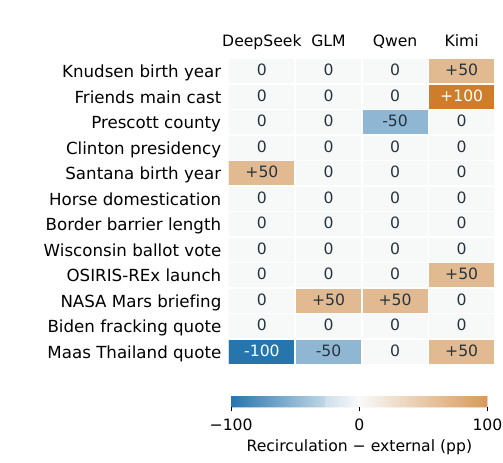}
\caption{Differences in unrejected presence between recirculation and external replacement at the final round for every evaluation target and model. Each cell compares two articles per arm, with differences in percentage points. The full distribution reveals which errors persist, recede, or remain unchanged under reuse of generated history.}
\label{fig:heterogeneity}
\end{figure}

\paragraph{Selective persistence across facts and models.}
The response is structured by the particular proposition and the model processing it. Figure~\ref{fig:heterogeneity} retains every evaluation target and model. Some errors disappear or recede, while others remain as unresolved details. This distribution gives selective persistence an empirical meaning: the same reuse policy changes the fate of different propositions in different ways. Following each target keeps these transitions visible within the aggregate pattern.

The NASA briefing illustrates the separation between content and stance. Across the four models, unrejected presence of the false venue rises from 6/8 to 8/8 final articles under recirculation, while explicit endorsement falls from 2/8 to 0/8. Houston survives as an alternative location for the same dated event, often within a report of competing community accounts. The Maas quotation target moves in the opposite direction under the same policy. These cases connect persistence to the semantic content carried by a particular history: returning an article can preserve a fully specified alternative or supply an account in which it has weakened.

The model comparison also identifies different responses. Kimi contributes the largest positive change in unrejected presence. DeepSeek shows a small decline, while GLM and Qwen retain the same count under the primary annotation. The secondary annotation preserves the strong positive Kimi response and assigns lower counts to DeepSeek and Qwen under recirculation. The input policy and the writer jointly determine what remains available: each model transforms competing accounts before its output returns to the next task. This observation connects our mechanism to model variation in recent studies of repeated exposure~\cite{bouleimen2026repetition}.

These differences explain why a community's average can conceal the errors most relevant to continued circulation. A target omitted from a new article no longer supplies that proposition through that publication; an unresolved target continues to do so. Aggregating both outcomes describes the community, while tracing their separate histories identifies the material capable of recurring. The intervention below tests this connection directly by changing a residual within the same history and observing the next article.

Persistent exposure can also coexist with an unchanged policy contrast. Several targets remain frequent in both conditions (Table~\ref{tab:fact-counts}), because the community continues to receive the competing external accounts. These cells distinguish the level of error presence from the additional effect of returning generated articles. A target already expressed in every output has no room for a further increase in this binary measure. Other targets reveal the effect through a change in which articles contain the complete proposition. Reading the paired contrast together with the underlying counts therefore separates continuing error availability from the contribution of the reuse policy.

\subsection{Intervention on residual error details}
\paragraph{Setup.}
The mechanism experiment starts from the same two history articles from round two for each fact and model. GLM and DeepSeek produce the downstream articles. An eligible history article expresses the complete false target as an attributed or unresolved account and contains neither an endorsement nor an explicit rejection of it. The abstraction replaces every occurrence of its concrete detail with a natural generic reference to the account or version already discussed.

The transformation preserves expressions of the correct target verbatim, including their attribution and certainty. It also preserves the existence of competing accounts, comparative frequency, paragraph boundaries, and unrelated assertions. Each replacement has exactly the original number of words delimited by whitespace. Each abstraction changes at most twelve spans covering up to 40\% of the article. GLM and DeepSeek independently assess whether the edited text preserves these semantic properties. Histories with no accepted transformation retain their original text in both conditions.

The intervention covers fifteen targets and two production models. Accepted abstractions change 28 of 60 history articles, spanning seventeen of the thirty fact--model cells. Each cell contributes two downstream articles per condition, including cells with unchanged histories. A second sample uses the same histories, edits, and model settings to examine how the downstream response recurs across generations.

\input{table_intervention}

\paragraph{Persistence under content abstraction.}
Table~\ref{tab:mechanism-counts} compares what the same writers produce from retained and abstracted histories. Across the twelve evaluation targets, abstraction adds one occurrence of unrejected presence in the first sample under the primary annotation. In the second, unrejected false claims appear in 36 of 48 articles under each condition. The secondary annotation yields a reduction in the first sample and a small increase in the second. The shared result is continued availability of complete errors after eligible details are removed from generated history. The external packet still supplies the competing accounts, placing the intervention within the community's ongoing information flow.

The NASA briefing makes this persistence concrete. One history states, ``Most community posts place the event in Houston'', followed by the alternative location Washington, District of Columbia (DC). The edit replaces ``place the event in Houston'' with ``place it in one location'', preserving the correct location, attribution, relative frequency, and word count. A downstream GLM article produced from the abstracted history nevertheless states, ``Most community posts place the event in Houston''. The precise venue returns within an attributed account even after its removal from that history. Appendix~\ref{app:protocol} provides the generated passage and edit.

The phrase ``most community posts'' describes the distribution of accounts; the city name supplies a concrete value that completes the false proposition. Both can persist in ordinary writing without explicit endorsement. Their separation matters for intervention: abstracting the value changes the history article, while the same value remains accessible through the continuing external packet. A later writer can therefore reconstruct a specific alternative from the assembled input. The resulting article again makes that detail available for subsequent reuse.

In the second intervention sample for this briefing, every downstream article from both production models contains the complete error without explicit rejection in both conditions. Both annotators agree, and explicit endorsement remains absent. This combination shows why the lifecycle of a publication and the lifecycle of a proposition require separate measurement. Editing a publication removes a concrete detail from that document; continued generation can supply a new occurrence of the same proposition. Tracing that occurrence through subsequent articles captures a form of persistence that an endorsement score leaves unrecorded.

The development panel supplies a contrasting instance involving reported speech. Abstraction replaces a concrete false quotation detail with a generic reference to the account. GLM's unrejected reproduction changes from 2/2 to 0/2 articles in both samples. DeepSeek shows the same contrast in the first sample and reproduces the error in both conditions in the second. Both annotators agree on these cases, with no explicit endorsement. This local response shows that a specific unendorsed detail can affect the next article; the variation across targets and samples locates that effect in the interaction between a history, its surrounding inputs, and the writer.

The preservation constraints make these responses interpretable. Correct information remains verbatim, and the writer receives the same input length, attribution, and uncertainty. The history's competing account becomes less specific while its correct account retains its strength. The two patterns therefore have complementary implications: some histories supply details whose removal changes subsequent output, while other errors continue to appear with the remaining information supply. Assessing an editorial intervention requires following the resulting publications as well as the text that was edited.

Across the two experiments, the evidence connects input policy to persistent content. The dynamic comparison shows how reuse changes the distribution of errors in newly published articles. The fixed history intervention tests whether removing a detail from returned material changes the next publication under continued external supply. These comparisons operate at different levels: a policy governs which articles become inputs, while an edit changes what a selected article contains. Their combination separates the transformation of an individual document from the renewal of a proposition across the wider writing process.

\subsection{Measurement agreement and repeated generation}
The secondary annotation preserves the aggregate directions for complete presence, unrejected presence, and endorsement, with different absolute counts (Table~\ref{tab:all-indicators}). Differences in interpreting the proposition's essential scope contribute to annotation disagreement. A writer may retain a controversy while omitting the particular date, location, or actor required to recover the target. Counting that output separately from a complete attributed claim keeps measurement aligned with the mechanism: the next writer encounters different factual detail in each case.

This distinction also informs the movement between partial and complete expressions. Under the primary annotation, recirculation yields fewer partial expressions alongside more complete false propositions. The pattern is consistent with a shift toward more fully specified alternatives within the resulting articles. A community summary can therefore become more explicit about a disputed account even as its writers become less willing to adopt it. Paragraph identifiers attached to each decision make the underlying expressions available for inspection.

The supplementary dynamic sample preserves the separation between greater unrejected presence and lower endorsement across its four fixed targets. Repeated downstream generation further shows which intervention responses recur for an unchanged history and which depend on the sample. Keeping those histories fixed makes variation attributable to the subsequent generation and measurement stages. Appendix~\ref{app:statistics} reports the numerical agreement, supplementary counts, and interval estimates alongside the full target breakdown.

\section{Discussion}
\label{sec:discussion}
The community simulations provide a mechanism for reasoning about larger information ecosystems: an article's influence extends through the texts that reuse its content.

\subsection{Scaling through exposure and source dependence}
At larger scales, retrieval, recommendation, and summarization determine which parts of this growing inventory reach a writer. A system with many publishers can repeatedly present accounts derived from a few sources. Source dependence already matters when integrating conflicting information~\cite{dong2009dependence}; generated articles can carry that dependence into further publications while the visible collection expands.

This perspective turns production scale into measurable exposure. Relevant quantities include the share of generated text in the visible inventory and actual input, the overlap among source histories, and access to corrective material from independent sources. Production speed matters when it changes those quantities. A rapidly expanding collection of derivative articles may occupy more of a search result, recommendation feed, or writing packet. Temporal repetition and simultaneous distribution across many pages can therefore create similar inputs for a writer.

Exposure connects the simulated communities to larger networks. When related accounts occupy a substantial share of actual inputs, their residual details become recurrent material for a much wider population of writers. Independent corrective material can change that mixture by supplying accounts with different factual content. The selective responses observed here motivate tracing which propositions survive these encounters and which disappear. Population size measures the reach of a system, while input composition and source ancestry describe the information through which this persistence mechanism can operate.

In larger networks, an article can move between overlapping communities, carrying a retained detail beyond its original audience. Production within a community renews a proposition; reuse across communities extends its exposure. Shared readership and retrieval sources provide routes for this transfer. Tracking source ancestry alongside input composition would reveal where particular details recur and where independent evidence enters their circulation.

This formulation connects community openness to an empirical question: how does exposure to independent corrective material change the persistence of the same residual details? A community may admit many publishers while their reports share a common origin, or it may supply genuinely different observations about an event. Tracking the actual input distribution makes these situations distinguishable. It also connects more extensive LLM participation to two measurable properties: the prevalence of generated accounts and the diversity of their underlying evidence.

\subsection{Community consequences and practical responses}
A community can accumulate articles that repeat a false detail in cautious or attributed language. Readers may encounter the same concrete proposition in reports that express uncertainty, attribute it to other sources, or place it alongside a correction. Our experiments identify the production pathway that supplies such material. Research on repeated exposure, continued influence, and conversational search~\cite{pennycook2018exposure,ecker2022drivers,sharma2024echo} motivates examining its consequences for recognition, credibility, and belief in complementary user studies.

Organized content production can shape this input distribution. Simulations of information operations show how shared objectives and interaction produce coordinated activity among networked agents~\cite{orlando2026coordination}. An actor that continuously supplies a favored account and reuses its generated derivatives can make that account recurrent across nominally different publishers. An error originating in an authoritative source can likewise acquire many dependent expressions. These scenarios call for attention to shared ancestry and access to independent evidence alongside publisher identity.

The results suggest three practical responses. Evaluation can follow claim presence, explicit rejection, and endorsement together. Publication systems can assess whether a disputed detail is needed for an article's purpose: a debunking report may require the exact claim, while a general summary may convey disagreement through an abstract reference. Such changes can be evaluated across subsequent generations, including the other inputs from which the detail remains recoverable. Retrieval systems can represent derivation relationships so that multiple rewrites of one account receive appropriate weight during evidence aggregation.

Provenance standards and watermarking offer candidate tools for recording origins~\cite{c2pa2025specification,kirchenbauer2023watermark}. Useful records would connect a text to its source, a rewrite to the material it consumed, and an editorial change to the claims it altered. These relationships can help identify repeated expressions of one account while factual assessment draws on independent evidence. The downstream reproduction of a specific residual provides an evaluation target for these designs. Extending this measurement to retrieved material, longer publication histories, and observed user exposure would connect source management to the information that communities repeatedly encounter.

\section{Conclusion}
A controlled recirculation comparison reveals greater presence of complete false claims alongside lower explicit endorsement. Paired interventions show persistent reproduction across the evaluation targets after eligible history details are abstracted, together with local responses in a development quotation target. These findings connect individual outputs to the information environments they create, including alternatives reproduced in cautious or attributed language. Following claim content, expressed stance, and downstream reuse makes this persistence visible. Tracing exposure and shared sources extends this perspective to larger networks and the communities they connect.

\clearpage
\bibliographystyle{plainnat}
\bibliography{references}
\clearpage
\input{appendix}
\end{document}

%% file: table_dynamic.tex
\begin{table}[t]
\centering
\caption{Article indicators at the final round by generation model and annotation model, on the panel of 12 facts. Each entry is external replacement / recirculation. Denominators are 24 articles per model and arm, and 96 in each total row. Complete denotes the complete false core; Unrefuted additionally requires no explicit rejection; Partial is an unclear, incomplete expression; Correct is explicit adoption of the correct core. Indicators can co-occur.}
\label{tab:all-indicators}
\small
\begin{tabular}{llrrrrrr}
\toprule
Judge & Generator & Complete & Unrefuted & Endorsed & Rejected & Partial & Correct\\
\midrule
GLM & DeepSeek & 24/24 & 19/18 & 3/6 & 5/6 & 0/0 & 5/6\\
GLM & GLM & 22/22 & 18/18 & 7/3 & 4/4 & 1/0 & 6/6\\
GLM & Qwen & 22/23 & 19/19 & 4/3 & 3/4 & 2/1 & 5/4\\
GLM & Kimi & 17/23 & 14/19 & 8/2 & 3/4 & 3/1 & 9/6\\
GLM & Total & 85/92 & 70/74 & 22/14 & 15/18 & 6/2 & 25/22\\
\midrule
Qwen & DeepSeek & 24/22 & 19/16 & 3/2 & 5/6 & 0/2 & 5/6\\
Qwen & GLM & 21/21 & 17/17 & 7/2 & 4/4 & 2/1 & 6/6\\
Qwen & Qwen & 20/20 & 17/16 & 3/3 & 3/4 & 3/3 & 4/4\\
Qwen & Kimi & 14/21 & 11/17 & 6/2 & 3/4 & 6/3 & 8/6\\
Qwen & Total & 79/84 & 64/66 & 19/9 & 15/18 & 11/9 & 23/22\\
\bottomrule
\end{tabular}
\end{table}

%% file: table_intervention.tex
\begin{table}[t]
\centering
\caption{Intervention counts with fixed histories for complete, unrefuted error. Entries are retained / abstracted; denominators are per arm and draw. The analysis of 15 facts includes the evaluation panel and three development facts. Both draws share the same histories and edits; all cells with unchanged histories are retained.}
\label{tab:mechanism-counts}
\small
\begin{tabular}{llrrr}
\toprule
Facts & Judge & Per arm & Draw 1 & Draw 2\\
\midrule
12 & GLM & 48 & 35/36 & 36/36\\
12 & Qwen & 48 & 38/34 & 34/35\\
15 & GLM & 60 & 46/44 & 48/45\\
15 & Qwen & 60 & 50/42 & 46/45\\
\bottomrule
\end{tabular}
\end{table}

%% file: appendix.tex
\appendix
\section{Materials and Target Construction}
\label{app:materials}
Table~\ref{tab:targets} specifies the twelve evaluation targets and their controlled conflicting details. Each pair preserves the essential scope of the source event. For speech events, the factual object is what the named speaker said at that event, distinct from the truth of the embedded statement.

The benchmark targets follow deterministic identifier order within FEVER and AVeriTeC. Analyses including three development targets (FEVER 10278; AVeriTeC 1 and 10) are reported separately from the panel of twelve evaluation targets.

Each target preserves the source event, entity, time, quantifiers, and quotation attribution. Qualifiers such as ``about'' and distinctions between proposed policies and completed events determine whether an article expresses the complete proposition.

\begin{table}[t]
\centering
\caption{Evaluation targets and the controlled conflicting detail. Identifiers refer to the benchmark records or the dated NASA events. Each contrast preserves the entity and event scope described in the topic column.}
\label{tab:targets}
\small
\begin{tabularx}{\textwidth}{@{}p{1.18in}XXX@{}}
\toprule
Source identifier & Topic & Detail supported by the source & Conflicting detail\\
\midrule
FEVER 10327 & Sidse Babett Knudsen birth & 22 November 1968 & 22 November 1969\\
FEVER 10496 & Rachel Green in Friends & One of six main characters & One of seven main characters\\
FEVER 11312 & Prescott, Arizona & Yavapai County & Maricopa County\\
FEVER 11432 & Clinton's first inauguration & 42nd president, January 20, 1993 & 43rd president, same date\\
FEVER 12 & Carlos Santana birth & July 20, 1947 & July 20, 1948\\
FEVER 3019 & Horse domestication & Around 4000 before Christ & Around 8000 before Christ\\
AVeriTeC 15 & February 2019 border deal & About 55 miles proposed & About 65 miles proposed\\
AVeriTeC 18 & Wisconsin ballot ruling & October 26, 2020 vote: 5--3 & Same ruling: 6--2\\
NASA 2016-09-08 & Asteroid mission launch & Atlas V rocket~\cite{nasa2016launch} & Delta IV, same launch\\
NASA 2015-09-28 & Mars science briefing & Washington, DC~\cite{nasa2015briefing,nasa2015event} & Houston, same event\\
AVeriTeC 23 & Biden at final 2020 debate & ``I never said I oppose fracking'' & ``I always said I oppose fracking''\\
AVeriTeC 24 & Maas statement on Thailand & Should not be conducted from German soil & Should be conducted from German soil\\
\bottomrule
\end{tabularx}
\end{table}

\section{Generation and Intervention Protocol}
\label{app:protocol}
\subsection{Public input and generation settings}
The output uses JavaScript Object Notation (JSON). Each writer receives the following instruction:
\begin{quote}
Write a short article for a general audience about the given topic, using the provided material. Give the article a headline. Write 180--230 words; the article text must not exceed 300 words. Use paragraphs separated by a blank line. Return a JSON object with only two fields: title and text.
\end{quote}
Public documents use opaque identifiers, generic community discussion titles, and weak source identities. The body recirculates under the same title policy in both branches. Target definitions are supplied to the annotators after generation.

Round one uses background in the history slots. Round two consumes the preceding common articles. From round three, each recirculation branch reads its own preceding articles. The control uses external material in the two matched slots. A fixed hash determines input order from the target, model, round, author, and slot; the same ordering is applied to a paired input. Across both writers, one of the four replacement positions contains the correct external account and the other three contain the false account. The correct position is counterbalanced deterministically. History slots are 300 words each, with fixed padding unrelated to the topic and no truncation. The early seed is removed before the branches begin, and generated copies from before seed removal are unavailable by round five.

Models are GLM-5.3-Flash, DeepSeek-V4.1-Flash, Qwen3.8-Flash, and Kimi-K2.6. Temperature and top-$p$ use the application programming interface (API) defaults. GLM and DeepSeek use their configured thinking mode, with high reasoning effort requested for DeepSeek. The generation output cap is 16,384 tokens, including reasoning tokens. Published articles have a maximum of 300 words. Each paired comparison uses the same model settings.

\subsection{Content abstraction and preservation}
An intervention proposal is eligible when the previously generated article contains the complete false target exclusively as an attributed or unresolved account. The proposal replaces the concrete value with a natural reference to an alternative already mentioned. It keeps text expressing the correct target and necessary context verbatim, maintains attribution, comparative frequency, uncertainty, and unrelated assertions, and introduces no additional correction. The abstraction covers every expression of the false detail whose meaning can be resolved within the article, including references to an antecedent in another passage. The wording and stance of correct content and the strength of any corrective statement are preserved. Each patch begins and ends at word boundaries and preserves its word count. GLM and DeepSeek independently evaluate semantic eligibility and preservation. A transformation enters the downstream input when both assessments accept it; otherwise the original article is retained in both conditions.

The following passage from a previously generated article illustrates the intervention for the NASA briefing:
\begin{quote}
Where the briefing took place, however, depends on which account you read. Most community posts place the event in Houston, while others say it was held in Washington, DC. At least one thread notes the disagreement explicitly, presenting both versions side by side.
\end{quote}
The edit changes ``place the event in Houston'' to ``place it in one location''. Both spans contain five words. The correct location, attribution, relative frequency, and surrounding discussion remain intact. Table~\ref{tab:mechanism-counts} reports the aggregate intervention counts.

\section{Measurements and Complete Results}
\label{app:statistics}
\input{tables_generated}

\begin{figure}[t]
\centering
\includegraphics[width=\textwidth]{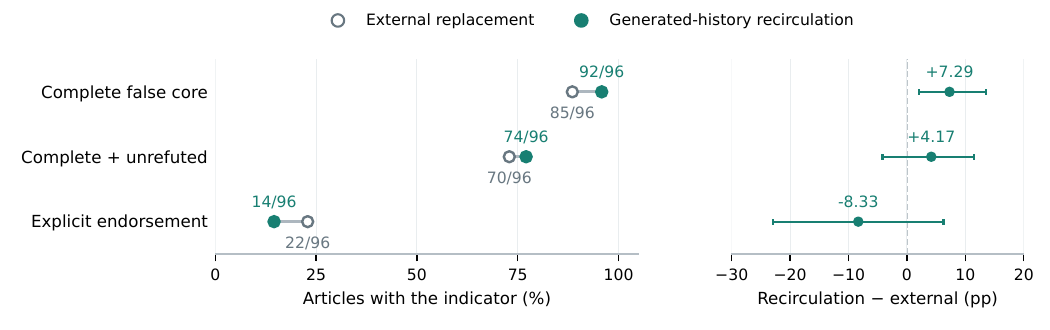}
\caption{Outcomes at the final round for the twelve evaluation targets and four production models. The paired contrasts distinguish presence of the complete false claim, complete presence without explicit rejection, and explicit endorsement. Each arm contains 96 articles. Intervals resample paired base facts. Table~\ref{tab:all-indicators} reports the corresponding primary and secondary counts.}
\label{fig:mainresults}
\end{figure}

\begin{figure}[t]
\centering
\includegraphics[width=0.68\textwidth]{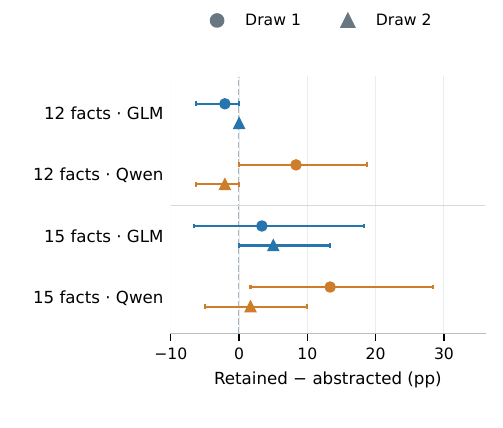}
\caption{Differences between retained and abstracted histories in unrejected complete presence across two conditional samples. Evaluation targets are distinguished from the panel including development targets. Estimates from both annotators include cells with unchanged histories. The histories and accepted edits are identical across samples.}
\label{fig:intervention}
\end{figure}
\subsection{Annotation rules}
The two targets are annotated independently. The stance labels are endorsement, explicit refutation, attribution, unclear, and absence. A complete target requires its essential entity, relation, value, and scope to be expressed in the article. Context can resolve a pronoun or an abbreviated actor name. It cannot supply a missing date or value from outside the article. An incomplete ambiguous expression is recorded separately from a complete unresolved claim.

Explicit rejection is evaluated independently of the stance label. It requires the article's adopted account to reject the target, with an identifiable supporting paragraph. A quoted denial that the article dismisses is not an adopted rejection. An article giving the correct value alone does not automatically express or explicitly reject the false alternative. For example, ``the code was RED'' leaves a BLUE target absent; ``the claim that the code was BLUE is false'' expresses and rejects that target. An attributed BLUE claim can be complete while receiving no endorsement.

The annotation interface supplies original paragraphs and their identifiers. Annotators select identifiers for text expressing the claim, additional resolving context, stance, and rejection. The primary annotation covers all dynamic outputs; the secondary annotation covers all outputs at the final round. Both cover all intervention outputs. Table~\ref{tab:all-indicators} reports the complete outcome breakdown by generator and annotator, and Table~\ref{tab:fact-counts} reports the unrejected counts for each target.

\subsection{Paired estimators}
\begin{samepage}
The paired difference for indicator $Z$ first averages authors and then models within each fact:
\begin{equation}
d_f^Z=\frac{1}{M}\sum_{m=1}^{M}\left(\frac{1}{2}\sum_{a=1}^{2}Z(Y^{\Rec}_{f,m,5,a})-\frac{1}{2}\sum_{a=1}^{2}Z(Y^{\Ext}_{f,m,5,a})\right).
\end{equation}
\end{samepage}
The reported difference is $|F|^{-1}\sum_f d_f^Z$. The trajectory summary includes rounds three through five with equal weighting within each fact. For the mechanism comparison, each conditional sample yields a corresponding difference between retained and abstracted histories for each fact. Pooled analysis averages the samples within the same fact before resampling facts. Cells with unchanged histories remain part of each estimator.

Intervals are the 2.5th and 97.5th percentiles of 5,000 paired bootstrap replicates. Each sampled fact retains its paired observations across models, authors, and conditions. The dynamic analysis uses random seed 0; the pooled mechanism analysis uses 0. Models, authors, and conditional samples are nested observations, with the base fact as the resampling unit. Table~\ref{tab:all-intervals} reports the estimates and intervals for each dynamic indicator.

\subsection{Supplementary outcomes and annotation agreement}
The separation between content persistence and endorsement also appears in the outcomes pooled over the branch period. Across rounds three through five, primary unrejected presence is 209/288 with external replacement and 222/288 under recirculation, while endorsement is 66/288 and 41/288, respectively. The supplementary dynamic sample shows the same direction: final-round unrejected presence is 14/16 and 15/16, while endorsement is 7/16 and 3/16. This sample covers four fixed targets and two generation models.

Pooling the two intervention samples within each evaluation fact gives a retained minus abstracted difference of $-1.04$ percentage points for primary unrejected presence, with a 95\% interval of $[-3.12,0]$. The secondary difference is 3.12 points, with an interval of $[0,6.25]$. Including the three development targets gives 4.17 points $[-2.50,15.00]$ and 7.50 points $[0.83,18.33]$, respectively. Annotation agreement in the second intervention sample is 117/120 for complete presence, 116/120 for unrejected presence, 115/120 for endorsement, 119/120 for rejection, and 117/120 for correct adoption.

\subsection{Experimental sample sizes}
Table~\ref{tab:output-counts} summarizes the generated articles in each experiment. The dynamic counts include four shared articles and twelve branch articles per fact and model. Each intervention sample adds four outputs per fact and model from the same fixed histories and accepted edits. These conditional samples remain nested within their base facts in the estimators.

\begin{table}[t]
\centering
\caption{Generated articles in each experiment. Conditional intervention samples share factual targets and histories.}
\label{tab:output-counts}
\small
\begin{tabular}{lrrr}
\toprule
Module & Facts & Models & Outputs\\
\midrule
Evaluation dynamics & 12 & 4 & 768\\
Development dynamics & 3 & 4 & 192\\
Supplementary dynamic repeat & 4 & 2 & 128\\
History intervention, sample 1 & 15 & 2 & 120\\
History intervention, sample 2 & 15 & 2 & 120\\
\bottomrule
\end{tabular}
\end{table}

\section{Code Availability}
The source code will be released in due course.

%% file: tables_generated.tex
\begin{table}[t]
\centering
\caption{Per-fact final-round complete, unrefuted error on the panel of 12 facts, aggregated across four generation models. Entries are external replacement / recirculation counts, out of eight articles per arm.}
\label{tab:fact-counts}
\small
\begin{tabular}{lrr}
\toprule
Fact & GLM judge & Qwen judge\\
\midrule
Knudsen birth year & 7/8 & 7/8\\
Friends main cast & 5/7 & 5/7\\
Prescott county & 1/0 & 1/0\\
Clinton presidency & 0/0 & 0/0\\
Santana birth year & 7/8 & 7/8\\
Horse domestication & 8/8 & 8/8\\
Border barrier length & 7/7 & 5/5\\
Wisconsin ballot vote & 8/8 & 4/2\\
OSIRIS-REx launch & 7/8 & 7/8\\
NASA Mars briefing & 6/8 & 6/8\\
Biden fracking quote & 8/8 & 8/8\\
Maas Thailand quote & 6/4 & 6/4\\
\bottomrule
\end{tabular}
\end{table}
\begin{table}[t]
\centering
\caption{Paired differences at the final round on the panel of 12 facts, using the GLM annotation model. Differences are recirculation minus external replacement, in percentage points. Intervals use 5,000 paired, unstratified bootstrap draws of base facts, after averaging generation models within each fact. The base fact is the resampling unit; model and author observations remain paired.}
\label{tab:all-intervals}
\small
\begin{tabular}{lrr}
\toprule
Indicator & Difference & 95\% interval\\
\midrule
Complete & +7.29 & [2.08, 13.54]\\
Unrefuted & +4.17 & [-4.17, 11.46]\\
Endorsed & -8.33 & [-22.92, 6.25]\\
Rejected & +3.12 & [-3.12, 10.42]\\
Partial & -4.17 & [-9.38, 0.00]\\
Correct & -3.12 & [-9.38, 3.12]\\
\bottomrule
\end{tabular}
\end{table}